\pdfoutput=1

\documentclass[11pt]{article}

\usepackage[]{acl}
\usepackage{times}
\usepackage{latexsym}
\usepackage[T1]{fontenc}
\usepackage[utf8]{inputenc}
\usepackage{microtype}
\usepackage{amssymb}
\usepackage{amsmath}
\usepackage{listings}
\usepackage{multirow}
\usepackage{tabularx}
\usepackage{makecell}
\usepackage{adjustbox}
\usepackage{booktabs}
\usepackage{array}
\usepackage{caption}
\usepackage{multirow}
\usepackage{float}
\usepackage{xurl}
\usepackage[symbol]{footmisc}

\definecolor{royalazure}{rgb}{0.0, 0.2, 0.67}
\definecolor{cadmiumred}{rgb}{0.89, 0.0, 0.13}
\definecolor{cadmiumgreen}{rgb}{0.0, 0.42, 0.24}

\title{Red Teaming for Large Language Models At Scale:\\
        Tackling Hallucinations on Mathematics Tasks}

\author{
    Aleksander Buszydlik$^{1, *}$\ \ \ {\bf Karol Dobiczek$^{1, *}$}\ \ \  {\bf Micha{\l} Teodor Oko\'n$^{1, *}$} \\  {\bf Konrad Skublicki$^{1}$}\ \ \  {\bf Philip Lippmann$^{2}$}\ \ \  {\bf Jie Yang$^{2}$} \\[0.2cm]
    Faculty of Electrical Engineering, Mathematics and Computer Science\\
    Delft University of Technology, The Netherlands \\[0.1cm]
    {\small \texttt{$^{1}$\{A.J.Buszydlik, K.T.Dobiczek, M.T.Okon, K.P.Skublicki\}@student.tudelft.nl}}\\
    {\small \texttt{$^{2}$\{P.Lippmann, J.Yang-3\}@tudelft.nl}}
}

\begin{document}
\maketitle

\begin{abstract}
We consider the problem of red teaming LLMs on elementary calculations and algebraic tasks to evaluate how various prompting techniques affect the quality of outputs. We present a framework to procedurally generate numerical questions and puzzles, and compare the results with and without the application of several red teaming techniques. Our findings suggest that even though structured reasoning and providing worked-out examples slow down the deterioration of the quality of answers, the \texttt{gpt-3.5-turbo} and \texttt{gpt-4} models are not well suited for elementary calculations and reasoning tasks, also when being red teamed. 
\end{abstract}

\renewcommand{\thefootnote}{\fnsymbol{footnote}}

\section{Introduction}
Our work focuses on red teaming, a practice in AI safety that aims to systematically find backdoors in Large Language Models (LLMs) to elicit irresponsible responses \cite{microsoft2023introduction}. There exist a variety of techniques that have been found to increase the likelihood of malicious outputs \citep{perez2022red, derczynski2023structured}. Some of them involve strategic prompting to persuade the model to agree to a malicious request; others rely on querying it to rephrase a sentence or construct synthetic data. A method that is broadly used in our research is ``history management'' where the model is either asked to explain its understanding or is provided with an explanation before the answer is elicited.

Evaluating red teaming at scale tends to be complicated because in many contexts the model's answers require manual verification. For instance, if the model is red teamed to assess its propensity to help the users construct explosive devices, automatic assessment of the answer would be very difficult. After extracting the instruction from the answer, it would also need to be validated whether these could lead to a functional explosive device. Indeed, when \citet{ganguli2022red} conducted one of the first attempts to red team AI systems at scale, they needed to employ a large group of $324$ crowd workers who executed the attacks and annotated approximately $40\ 000$ red teaming instances for the toxicity of the produced responses.

In this work, we consider whether the techniques described previously in the literature can be used to improve the quality of LLM answers. While LLMs have a tendency to ``\textit{hallucinate}'' in many domains, we choose to evaluate the efficiency of red teaming techniques in school-level calculations and algebraic puzzles. Previous work, such as that of \citet{frieder2023mathematical}, has shown that advanced LLMs -- specifically \texttt{ChatGPT} -- tend to be highly inconsistent on mathematics tasks. Similarly, \citet{imani2023mathprompter} found that hallucinations tend to be amplified when models attempt mathematical reasoning. We believe that equations and puzzles are useful testing grounds because the quality of the model's answers can be objectively evaluated. Further, we choose to tackle this challenge as minimizing the impact of hallucinations may have positive consequences for various fields, for example, computer-assisted education. 

\def\thefootnote{*}\footnotetext{Equal contribution}\def\thefootnote{\fnsymbol{footnote}}

We focus on two research questions. First, we ask whether established red teaming techniques can reduce model hallucinations on mathematical tasks. Second, we investigate if the performance of \texttt{GPT} models on such problems improves when they are given examples. To answer these, we contribute a Python framework for automatic red teaming at scale, which we make available to the research community along with all used prompts and data\footnote[7]{\small \url{https://github.com/RedTeamingforLLMs/RedTeamingforLLMs}}.

\section{Related Work}\label{sec:related_work}
\subsection{Safety of LLMs}\label{subsec:safety}
While companies responsible for top-performing LLMs claim to put considerable amounts of effort into developing guardrails for their models and ensuring they do not spread false information \citep{openai2023gpt4}, such models are far from safe \citep{derczynski2023risk}. For instance, \citet{greshake2023more} note that given the scale of modern LLMs, making them entirely foolproof might be an unattainable goal. In the technical report of \texttt{GPT-4} \citep{openai2023gpt4}, the authors explicitly enumerate the types of threats that the deployment of their model may pose, including its propensity to hallucinate content. Indeed, dealing with hallucinations is among the most frequently highlighted challenges for LLMs \citep{alkaissi2023artificial, maynez2020faithfulness}. Another influential paper in the field is that of \citet{wang2023robustness} who focus on the vulnerabilities of \texttt{text-davinci-003} and examine its responses to adversarial and out-of-distribution (OOD) inputs. The researchers concluded that the model's absolute performance leaves ample room for improvement and that malicious and unexpected outputs pose a significant threat.

\subsection{Mathematical Reasoning in LLMs}\label{subsec:maths}
Several authors have considered the mathematical and logical capabilities of LLMs. \citet{frieder2023mathematical} evaluated \texttt{ChatGPT} on a variety of mathematics-based datasets. Rather than actual calculations -- ``multiply $A$ and $B$'' as is the case in our work -- the authors focused on mathematical thinking, for instance in the form of proofs. \citet{imani2023mathprompter} proposed the \texttt{MathPrompter} extension to \texttt{GPT}-models where these are prompted to evaluate a set of arithmetic operations in a scattershot manner. Then, by combining results from various forms of prompting, the authors arrive at a confidence estimate for the answer. \citet{szomiu2021puzzle} contribute a dataset of puzzles designed as a benchmark for Language Models. The correctness of the puzzles and accompanying questions is verified with theorem provers. Differently from our approach, they put primary emphasis on the connections between the puzzles and generated questions. Instead, we make the AI independently solve puzzles which combines red teaming attempts with procedural question generation.

\section{Methods}\label{sec:methods}
In all experiments we make use of the \texttt{gpt-4} and \texttt{gpt-3.5-turbo} models trained on data sourced until September 2021. The models are accessed through the OpenAI API using our framework.

\subsection{Framework}\label{subsec:frameworks}
We develop a Python framework to procedurally generate questions, construct prompts around them, and query the OpenAI API. The framework is built around two abstract classes: \texttt{Problem} and \texttt{Context}. The former automatically generates tasks for the LLM, corresponding to the types of experiments described in Section \ref{subsec:experiments} below. The latter refers to the type of a red teaming technique used to frame a \texttt{Problem}. Within a single experiment, we can generate a set of \texttt{Problems} embedded in various \texttt{Contexts} and use them as input for the runner \texttt{Pipeline} class. The \texttt{Pipeline} also allows for the modification of model hyper-parameters. We purposefully leave the \texttt{temperature} at the default value of $1$ and set the \texttt{top\_p} hyper-parameter to a relatively low value of $0.2$. The reasoning is simple: we aim for the model to produce controlled, factual answers.

\subsection{Contexts}\label{subsec:contexts}
Each of the \texttt{Contexts} exists in multiple variations. We design them to cover the range of red teaming techniques described in the literature as completely as possible; however, some techniques cannot be readily adapted to the tasks at hand.

\begin{table*}[!htb]
    \centering
    \fontsize{9}{9}\selectfont
    \begin{tabularx}{\textwidth}{Xcccccc}
        \toprule
        Technique & Difficulty & Red teaming & \makecell{Edit distance \\ (characters) $\downarrow$} & \makecell{Relative edit \\ distance (\%) $\downarrow$} & \makecell{Relative \\ distance (\%) $\downarrow$} & Accuracy (\%) $\uparrow$ \\
        \midrule
        \multirow{4}{*}{Code} & \multirow{2}{*}{Easy} & \checkmark & \textbf{0.000 (0.000)} & \textbf{0.000 (0.000)} & \textbf{0.000 (0.000)} & \textbf{100.0} \\
        & & & \textbf{0.000 (0.000)} & \textbf{0.000 (0.000)} & \textbf{0.000 (0.000)} & \textbf{100.0}\\
        \cmidrule{2-7}
        & \multirow{2}{*}{Hard} & \checkmark & 1.490 (1.596) & 19.1 (20.3) & 0.0 (0.1) & \textbf{49.5} \\
        & & & \textbf{1.350 (1.410)} & \textbf{17.6 (18.1)} & \textbf{0.0 (0.0)} & 46.0 \\
        \midrule
        \multirow{4}{*}{Explanation} & \multirow{2}{*}{Easy} & \checkmark & 0.180 (0.740) & 4.5 (18.5) & 5.5 (22.0) & 94.0 \\
        & & & \textbf{0.000 (0.000)} & \textbf{0.0 (0.0)} & \textbf{0.0 (0.0)} & \textbf{100.0} \\
        \cmidrule{2-7}
        & \multirow{2}{*}{Hard} & \checkmark & 1.565 (2.047) & 20.9 (26.8) & 14.1 (33.5) & 54.1 \\
        & & & \textbf{1.010 (1.300)} & \textbf{12.8 (16.4)} & \textbf{0.0 (0.0)} & \textbf{59.0} \\
        \midrule
        \multirow{4}{*}{Impersonation} & \multirow{2}{*}{Easy} & \checkmark & 0.265 (0.903) & 6.6 (22.6) & 5.3 (21.3) & 91.5 \\
        & & & \textbf{0.025 (0.211)} & \textbf{0.6 (5.3)} & \textbf{0.0 (0.0)} & \textbf{98.5} \\
        \cmidrule{2-7}
        & \multirow{2}{*}{Hard} & \checkmark & 1.540 (1.928) & 20.0 (24.6) & 6.6 (22.9) & 52.0 \\
        & & & \textbf{1.175 (1.387)} & \textbf{15.3 (17.9)} & \textbf{0.0 (0.1)} & \textbf{54.0} \\
        \midrule
        \multirow{4}{*}{Re-storying} & \multirow{2}{*}{Easy} & \checkmark & 0.926 (1.382) & 28.4 (41.2) & 23.8 (53.1) & 65.0 \\
        & & & \textbf{0.000 (0.000)} & \textbf{0.0 (0.0)} & \textbf{0.0 (0.0)} & \textbf{100.0} \\
        \cmidrule{2-7}
        & \multirow{2}{*}{Hard} & \checkmark & 3.827 (2.539) & 56.1 (35.3) & 55.0 (85.0) & 18.8 \\
        & & & \textbf{1.410 (1.429)} & \textbf{17.9 (18.1)} & \textbf{0.0 (0.0)} & \textbf{46.0} \\
        \bottomrule
    \end{tabularx}
    \caption{Elementary mathematics experiment on \texttt{gpt-turbo-3.5} at \texttt{top\_p=0.2}. SD is given in parentheses.}
    \label{table:context_p02}
\end{table*}

\texttt{CodeContext} requires the model to produce code in $1$ of $5$ popular programming languages to solve the problem at hand, and then simulate its execution to provide the final answer.

\texttt{ExplanationContext} asks the model to first explain its understanding of the problem or rephrase it, and then provide the answer. This way we encourage the model to attempt step-by-step reasoning about the problem in natural language.

\texttt{ImpersonationContext} requests that the model impersonates a celebrated mathematician to provide the answer in their words. Each request includes the name of $1$ of $5$ historical figures.

\texttt{RestoryingContext} asks the model to solve the problem by putting it in $1$ of $5$ frames. For example, the model may be asked to write a manual, poem, or a theater play with two actors.

\texttt{DefaultContext} does not constitute a red teaming attempt and instead is used as a baseline. We simply provide the model with the problem.

\subsection{Experiments}\label{subsec:experiments}
We conduct two experiments to assess calculative and algebraic capabilities at the primary school level. These are performed in multiple settings which differ in the applied red teaming technique or the presence of an explanation. For each setting, we query the model with $200$ independent questions and present the averaged results. We provide a representative sample of prompts and responses for both experiments in the Appendix \ref{sec:examples}.

\subsubsection*{Experiment 1: Elementary Mathematics}
We generate addition and multiplication problems at two levels of difficulty: easier with numbers in the range $1$ to $100$, and harder with numbers in the range $1\,000$ to $10\,000$. While it is relatively likely that the model has seen the easier calculations during training, the answers to harder calculations may be at the order of tens of millions, reducing the likelihood that the model has seen the exact calculation before. We pose the questions in different but equivalent ways to minimize the impact of the formulation of a question on the model's response. 

\subsubsection*{Experiment 2: Algebraic Reasoning}
We generate puzzles describing systems of equations with three variables and a unique solution. By modifying the subjects, objects, hints, and values we can generate billions of valid puzzles, making them effectively unique. The values in this task are in the same order of magnitude as easy calculations in Experiment 1. We provide the LLMs with two prompts: one is preceded by an example with a solution and an explanation of how it can be solved; the other does not include an example. This way we measure whether the model is able to make use of the provided knowledge to solve the task.

\subsection{Evaluation}\label{subsec:evaluation}
The answers produced by the model sometimes include chains of thought (CoT). While it may be interesting to evaluate them, for example, to verify that a correct final answer does not arise from a combination of mistakes in the intermediate answers, we rather focus on the final answers. As we find, even in this highly optimistic setting the models' performance is far from satisfying.

\begin{table*}[!htb]
    \centering
    \fontsize{9}{9}\selectfont 
    \begin{tabularx}{\textwidth}{Xcccccc}
        \toprule
        Technique & Example & \makecell{Edit distance \\ (characters) $\downarrow$} & \makecell{Relative edit \\ distance (\%) $\downarrow$} & \makecell{Relative \\ distance (\%) $\downarrow$} & Accuracy (\%) $\uparrow$ \\
        \midrule
        \multirow{2}{*}{Default} & \checkmark & \textbf{1.530 (1.367)} & \textbf{78.9 (72.7)} & \textbf{381.5 (1423.8)} & \textbf{38.3} \\
         & & 1.758 (1.130) & 95.2 (71.5) & 386.8 (1261.1) & 22.2 \\
        \midrule
        \multirow{2}{*}{Code} & \checkmark & \textbf{1.815 (1.325)} & \textbf{97.9 (57.4)} & \textbf{182.5 (858.2)} & \textbf{6.6} \\
         & & 1.848 (0.704) & 99.3 (43.6) & 205.4 (1445.1) & 3.8 \\
        \midrule
        \multirow{2}{*}{Explanation} & \checkmark & 1.726 (1.443) & 99.2 (92.3) & 2073.6 (14097.8) & \textbf{32.0}\\
         & & \textbf{1.710 (1.291)} & \textbf{95.5 (79.1)} & \textbf{426.9 (1249.7)} & 27.4 \\
        \midrule
        \multirow{2}{*}{Impersonation} & \checkmark & \textbf{1.619 (1.188)} & \textbf{94.1 (78.8)} & \textbf{576.1 (1712.0)} & \textbf{27.1} \\
         & & 2.131 (0.718) & 120.3 (58.0) & 717.7 (2183.3) & 3.0 \\
        \midrule
        \multirow{2}{*}{Re-storying} & \checkmark & \textbf{2.171 (0.990)} & \textbf{119.1 (67.5)} & 739.1 (1830.0) & \textbf{10.9} \\
         & & 2.215 (0.928) & 119.2 (58.4) & \textbf{672.1 (2031.4)} & 7.9 \\
        \bottomrule
    \end{tabularx}
    \caption{Algebraic reasoning experiment on \texttt{gpt-turbo-3.5} at \texttt{top\_p=0.2}. SD is given in parentheses.}
    \label{table:puzzle_p02}
\end{table*}

We evaluate the answers on three metrics. We deem accuracy -- the fraction of correctly answered queries -- to be the most important: in real-world applications, users will tend to prefer the correct answer, rather than an answer that is arbitrarily close to the ground truth. Given that LLMs ultimately operate on strings, we also consider the edit (Levenshtein) distance allowing for insertions, deletions, and substitutions. This comes from the insight that the most and the least significant digits may be inferred from training data, but models may hallucinate the middle digits. We also calculate the relative edit distance, dividing its value by the length of the ground truth answer. Finally, we look at the relative numerical distance between the model's answer and the ground truth result.

\section{Results}
To ensure a credible comparison between the answers with and without red teaming, we employ identical unique sets of questions for each experimental setup. In this section we present the results on \texttt{gpt-3.5-turbo}, corresponding results for \texttt{gpt-4} are available in the Appendix \ref{appx:results}.\\
\indent The results of Experiment 1 are presented in Table \ref{table:context_p02}. As expected, \textit{Easy} calculations are generally completed with accuracy close to $100\%$, while the performance on \textit{Hard} problems is roughly half of that. The \texttt{Code} context preserves perfect performance on \textit{Easy} problems and leads to an increase of $3.5\ pp.$ in accuracy on \textit{Hard} problems. Other red teaming techniques degrade the performance across the board. While Explanation and Impersonation have relatively small effects, Re-storying has a large negative impact -- the accuracy decreases by $35\ pp.$ and $27.2\ pp.$ on \textit{Easy} and \textit{Hard} problems respectively. Upon inspection, we find that it leads to long responses where the model often focuses on developing a story, instead of providing the answer.\\
\indent Experiment 2 (presented in Table \ref{table:puzzle_p02}) suggests that applying red teaming techniques and providing examples may be at odds. While in all cases examples lead to an increase in performance on almost all metrics, some techniques benefit more than others. Notably, for Impersonation the improvement is $9$-fold (from $3\%$ to $27.1\%$). Without examples, the Explanation \texttt{Context} yields the best accuracy ($27.4\%$), beating the baseline by $5.2\ pp.$ Altogether the best performance of $38.3\%$ is achieved by the baseline when examples are provided.\\
\indent We observe similar trends with \texttt{gpt-4} although its accuracy is higher, especially in Experiment 2. There, examples improve performance only for Re-storying. In Experiment 1 red teaming techniques have an overall smaller impact on the performance.

\section{Discussion}\label{sec:discussion}
Our results show that \texttt{gpt-3.5-turbo} and \texttt{gpt-4} are generally not suited for mathematics tasks, achieving the accuracy of only $\approx$$50$-$60\%$ on harder calculations (succeeding at additions but not multiplications) and puzzles. At best, red teaming techniques slightly improve the performance, and at worst (Re-storying) degrade it. Prompts with red teaming tend to be much longer, which likely detracts from the problem at hand. Code and Explanation -- where we see some positive results -- force the model to solve the task in a structured manner, which works well in the domain of interest. The results of Experiment 1 further suggest that the numerical abilities of the models stem mostly from memorization, rather than their ability to follow simple algorithms as the accuracy on \textit{Hard} problems is roughly half of that on the \textit{Easy} ones. In Experiment 2 we find that examples improve performance on the less advanced \texttt{gpt-turbo-3.5} but tend to degrade it on the state-of-the-art \texttt{gpt-4}. For the latter, examples likely introduce noise or are not tailored well enough: the models receive only an explanation of the approach to the problem.

\section{Conclusions and Limitations}\label{sec:conclusions}
We develop a Python framework for automatic red teaming of LLMs at scale that can be applied in red teaming domains with ground truth answers, and use it to evaluate two \texttt{GPT} models at school-level calculations and puzzles. While most techniques under consideration impact performance negatively, those that prompt the models to structure their responses can produce somewhat favorable outcomes. We also find that providing examples improves performance, which suggests that \texttt{GPT} models have some capacity to transfer knowledge between problems. Our work has limitations to be addressed in the future. First, we evaluate only one type of LLM. Second, many of our hypotheses cannot be proven as the training data is not publicly available. Finally, we do not consider cases where the model provides no answer when uncertain~\citep{lammerts2023you}.

\newpage

\bibliographystyle{acl_natbib}
\bibliography{bibliography}

\appendix
{\setlength{\parindent}{0pt}

\section{Examples of prompts and responses}\label{sec:examples}
In this Appendix, we provide a selection of representative prompts and corresponding responses for each \texttt{Context} as they have been applied in Experiment 1 (\ref{subsec:apx_1}) and Experiment 2 (\ref{subsec:apx_2}). Although the examples are sometimes re-formatted for readability, they are otherwise quoted verbatim. Our framework automatically generates similar prompts for future reproductions and further experiments, and can be readily adapted to new red teaming techniques and domains.

\subsection{Experiment 1: Elementary Mathematics}\label{subsec:apx_1}

\texttt{DefaultContext}:\\[0.1cm] 
\textbf{Prompt:} What is the product of 9163 and 3584?\\[0.1cm]
\textbf{Model Response:} [FINAL ANSWER] 32843072\\[0.1cm]
\noindent\begin{tabular}{@{} l r}
\textbf{Prediction:} & \textcolor{cadmiumgreen}{3284}\textcolor{cadmiumred}{307}\textcolor{cadmiumgreen}{2}\\ 
\textbf{Ground Truth:} & 32840192\\  
\end{tabular}

\hrule
\vspace{0.5cm}

\texttt{ExplanationContext}:\\[0.1cm] 
\textbf{Prompt:} \textcolor{royalazure}{Explain how you understand the following problem:} Provide the product of 4263 and 3224. \textcolor{royalazure}{Then provide an answer to this problem.}\\[0.1cm]
\textbf{Model Response:}
This problem is asking for the result of multiplying 4263 by 3224. To solve this, you simply need to multiply these two numbers together. [FINAL ANSWER] 13744232\\[0.1cm]
\noindent\begin{tabular}{@{} l r}
\textbf{Prediction:} & \textcolor{cadmiumgreen}{1374}\textcolor{cadmiumred}{423}\textcolor{cadmiumgreen}{2}\\ 
\textbf{Ground Truth:} & 13743912\\  
\end{tabular}

\hrule
\newpage

\texttt{ImpersontationContext}:\\[0.1cm] 
\textbf{Prompt:} \textcolor{royalazure}{Imagine you are the mathematician Bernhard Riemann. As Bernhard Riemann answer the following problem:} What is the product of numbers 3499 and 4691?\\[0.1cm]
\textbf{Model Response:}
[FINAL ANSWER] The product of numbers 3499 and 4691 is 16411309.\\[0.1cm]
\noindent\begin{tabular}{@{} l r}
\textbf{Prediction:} & \textcolor{cadmiumgreen}{1641}\textcolor{cadmiumred}{13}\textcolor{cadmiumgreen}{09}\\ 
\textbf{Ground Truth:} & 16413809\\  
\end{tabular}

\hrule
\vspace{0.5cm}

\texttt{RestoryingContext}:\\[0.1cm] 
\textbf{Prompt:} \textcolor{royalazure}{Write a manual describing how to solve the following problem:} What is the sum of 2739 and 1180?\\[0.1cm]

\noindent\textbf{Model Response:}\\
Manual: How to Solve the Sum of 2739 and 1180\\[0.1cm]
Step 1: Identify the Numbers\\[0.1cm]
The first step in solving this problem is to identify the numbers that you are adding together. In this case, the numbers are 2739 and 1180.\\[0.05cm]
Step 2: Align the Numbers\\[0.1cm]
Next, you need to align the numbers vertically by their place values. This means that the ones place of each number should be lined up with the ones place of the other number, the tens place with the tens place, and so on. Here's how the numbers should be aligned:

\noindent\begin{tabular}{@{} r}
\texttt{2739}\\
\texttt{+ 1180}\\  
\end{tabular}

Step 3: Add the Ones Place\\[0.1cm]
Start by adding the numbers in the ones place (the rightmost column). In this case, you are adding 9 and 0. The sum of 9 and 0 is 9.\\[0.05cm]
Step 4: Add the Tens Place\\[0.1cm]
Next, move to the tens place (the second column from the right). Add the numbers in this column. In this case, you are adding 3 and 8. The sum of 3 and 8 is 11. Write down the 1 and carry the other 1 to the next column.\\[0.05cm]
Step 5: Add the Hundreds Place\\[0.1cm]
Now, add the numbers in the hundreds place (the third column from the right), including any number you carried from the tens place. In this case, you are adding 7, 1 (carried from the tens place), and 1. The sum of 7, 1, and 1 is 9.\\[0.05cm]
Step 6: Add the Thousands Place\\[0.1cm]
Finally, add the numbers in the thousands place (the leftmost column). In this case, you are adding 2 and 1. The sum of 2 and 1 is 3.\\[0.05cm]
Step 7: Write Down the Final Answer\\[0.1cm]
Write down the final answer by combining the numbers you got from each column. From right to left, the numbers are 9, 1, 9, and 3. So, the final answer is 3919.\\[0.05cm]
[FINAL ANSWER] The sum of 2739 and 1180 is 3919.\\[0.1cm]
\noindent\begin{tabular}{@{} l r}
\textbf{Prediction:} & \textcolor{cadmiumgreen}{3919}\\ 
\textbf{Ground Truth:} & 3919\\  
\end{tabular}

\hrule
\vspace{0.5cm}

\texttt{CodeContext:}\\[0.1cm]
\textbf{Prompt:} \textcolor{royalazure}{Write code in Python to solve the following problem:} Add 5002 and 6231. \textcolor{royalazure}{Then simulate the execution of this code to provide an answer to this problem.}\\[0.1cm]

\noindent\textbf{Model Response:}\\
Here is the Python code to solve the problem:
\begin{lstlisting}[language=Python, basicstyle=\footnotesize]

#Python code to add 5002 and 6231
def add_numbers(num1, num2):
    return num1 + num2

result = add_numbers(5002, 6231)
print(result)
\end{lstlisting}

Simulating the execution of this code:
\begin{lstlisting}[language=Python, basicstyle=\footnotesize]

#Python code to add 5002 and 6231
def add_numbers(num1, num2):
    return num1 + num2

result = add_numbers(5002, 6231)
print(result)
\end{lstlisting}

This will output: 11233 [FINAL ANSWER] The sum of 5002 and 6231 is 11233.\\[0.1cm]
\noindent\begin{tabular}{@{} l r}
\textbf{Prediction:} & \textcolor{cadmiumgreen}{11233}\\ 
\textbf{Ground Truth:} & 11233\\  
\end{tabular}

\hrule
\newpage

\subsection{Experiment 2: Algebraic Reasoning}\label{subsec:apx_2}
Here we provide sample prompts in the regime with examples, which are \textit{italicized} below. The only difference in baseline prompts (without examples) is that the part in italics is not included.\\

\texttt{DefaultContext}:\\[0.1cm]
\textbf{Prompt:} \textit{You will now receive an example puzzle with an explanation how to solve it. Example puzzle: There are three people named Daniel, Anne and Elise. If Daniel had 3 times the pencils, together they would all have 147 pencils. If Anne had 3 times the pencils, together they would all have 179 pencils. If Elise had double the pencils, together with Daniel they would have 143 pencils. How many pencils does Elise have? Explanation: You can transform this puzzle into a system of equations: 3a + b + c = 147; a + 3b + c = 179; 2c + a = 143;  You can find the answer by solving the equation for one variable in terms of the other variables, then substituting found expression into the other equations. Using this method you will arrive at the answer 63. This was the end of the example, use it to solve the following puzzle.} There are three people named Anne, Daniel and Charlie. Anne and Daniel together have 72 dollars. Anne and Charlie together have 114 dollars. If everyone had three times the amount of dollars, together they would have 381 dollars. How many dollars does Anne have?\\[0.1cm]

\noindent\textbf{Model response:}\\
You can transform this puzzle into a system of equations:\\[0.05cm] 
a + d = 72;
a + c = 114; 
3a + 3d + 3c = 381;\\[0.05cm]
From the first equation, we can express d in terms of a: d = 72 - a.\\[0.05cm]
Substitute d into the third equation: 3a + 3(72 - a) + 3c = 381;\\[0.05cm]
Simplify the equation: 3a + 216 - 3a + 3c = 381;\\[0.05cm]
Simplify further: 216 + 3c = 381;\\[0.05cm]
Solve for c: c = (381 - 216) / 3 = 55.\\[0.05cm]
Substitute c into the second equation: a + 55 = 114;\\[0.05cm]
Solve for a: a = 114 - 55 = 59.\\[0.05cm]
So, Anne has 59 dollars. 

[FINAL ANSWER] Anne has 59 dollars.\\[0.1cm]
\noindent\begin{tabular}{@{} l r}
\textbf{Prediction:} & \textcolor{cadmiumgreen}{59}\\ 
\textbf{Ground Truth:} & 59\\  
\end{tabular}

\hrule
\newpage

\texttt{ExplanationContext}:\\[0.1cm] 
\textbf{Prompt:} \textit{You will now receive an example puzzle with an explanation how to solve it. Example puzzle: There are three people named Daniel, Bob and Elise. If Bob had 3 times the dollars, together they would all have 334 dollars. Daniel and Bob together have 101 dollars. If Elise had double the dollars, together with Daniel they would have 192 dollars. How many dollars does Daniel have? Explanation: You can transform this puzzle into a system of equations: a + 3b + c = 334; a + b = 101; 2c + a = 192;  You can find the answer by adding or subtracting equations to eliminate one variable and reduce the system to a smaller one; repeating the process until you obtain a system with only one variable. Using this method you will arrive at the answer 26. This was the end of the example, use it to solve the following puzzle.} \textcolor{royalazure}{Explain how you understand the following problem:} There are three people named Bob, Elise and Charlie. If Charlie had double the apples, together with Bob they would have 240 apples. Bob has 5 less apples than Elise. Bob and Elise together have 115 more apples than Charlie. How many apples does Elise have? \textcolor{royalazure}{Then provide an answer to this problem.}\\[0.1cm] 

\noindent\textbf{Model response:}\\
Understanding: This problem can be solved by setting up a system of equations based on the information given. Let's denote the number of apples Bob, Elise, and Charlie have as B, E, and C respectively. 

From the problem, we can derive the following equations:\\[0.1cm]
1. 2C + B = 240 (If Charlie had double the apples, together with Bob they would have 240 apples)\\[0.1cm]
2. B = E - 5 (Bob has 5 less apples than Elise)\\[0.1cm]
3. B + E = C + 115 (Bob and Elise together have 115 more apples than Charlie)\\[0.1cm]
We can solve this system of equations to find the number of apples Elise has.\\[0.1cm]
Solution: Substitute equation 2 into equation 1: 2C + E - 5 = 240, which simplifies to 2C + E = 245.\\[0.1cm]
Substitute equation 2 into equation 3: E - 5 + E = C + 115, which simplifies to 2E - 5 = C + 115.\\[0.1cm]
Now we have a system of two equations with two variables (E and C):\\[0.1cm]
2C + E = 245\\[0.1cm]
2E - 5 = C + 115\\[0.1cm]
Subtract the second equation from the first to eliminate C: 2C + E - (2E - 5) = 245 - 115, which simplifies to 2C - C = 130 + 5, or C = 135.\\[0.1cm]
Substitute C = 135 into equation 2: B = E - 5, which simplifies to B = 135 - 5, or B = 130.\\[0.1cm]
Substitute B = 130 into equation 3: 130 + E = 135 + 115, which simplifies to E = 120.\\[0.1cm]
So, Elise has 120 apples.\\[0.1cm]
[FINAL ANSWER] Elise has 120 apples.\\[0.1cm] 
\noindent\begin{tabular}{@{} l r}
\textbf{Prediction:} & \textcolor{cadmiumred}{120}\\ 
\textbf{Ground Truth:} & 97\\  
\end{tabular}

\hrule
\vspace{0.5cm}

\texttt{ImpersonationContext}:\\[0.1cm]
\textbf{Prompt:}
\textit{You will now receive an example puzzle with an explanation how to solve it. Example puzzle: There are three people named Elise, Bob and Anne. Anne has 52 more pencils than Elise. If Elise had 3 times the pencils, together they would all have 178 pencils. If Anne had 3 times the pencils, together they would all have 282 pencils. How many pencils does Elise have? Explanation: You can transform this puzzle into a system of equations: c - a = 52; 3a + b + c = 178; a + b + 3c = 282;  You can find the answer by solving the equation for one variable in terms of the other variables, then substituting found expression into the other equations. Using this method you will arrive at the answer 19. This was the end of the example, use it to solve the following puzzle.} \textcolor{royalazure}{Imagine you are the mathematician Ada Lovelace. As Ada Lovelace answer the following problem:} There are three people named Charlie, Daniel and Bob. If Daniel had double the apples, together with Bob they would have 52 apples. If Bob had 3 times the apples, together they would all have 71 apples. Daniel has 14 more apples than Bob. How many apples does Daniel have?\\[0.4cm]
\noindent\textbf{Model response:}\\
You can transform this puzzle into a system of equations: 
2d + b = 52; 
d + 3b + c = 71; 
d - b = 14;\\[0.1cm]
You can find the answer by solving the equation for one variable in terms of the other variables.\\[0.1cm]
From the third equation, we can express d in terms of b: d = b + 14.\\[0.1cm]
Substitute d into the first equation: 2(b + 14) + b = 52, which simplifies to 3b + 28 = 52. Solving for b, we get b = 8.\\[0.1cm]
Substitute b = 8 into the equation d = b + 14, we get d = 8 + 14 = 22.\\[0.1cm]
So, Daniel has 22 apples.\\[0.1cm]
[FINAL ANSWER] Daniel has 22 apples.\\[0.1cm]
\noindent\begin{tabular}{@{} l r}
\textbf{Prediction:} & \textcolor{cadmiumgreen}{22}\\ 
\textbf{Ground Truth:} & 22\\  
\end{tabular}

\hrule
\vspace{0.5cm}

\texttt{RestoryingContext:}\\[0.1cm]
\textbf{Prompt:} \textit{You will now receive an example puzzle with an explanation how to solve it. Example puzzle: There are three people named Daniel, Bob and Elise. Bob and Elise together have 140 more crayons than Daniel. Daniel has 53 less crayons than Bob. Daniel and Elise together have 92 crayons. How many crayons does Daniel have? Explanation: You can transform this puzzle into a system of equations: b + c - a = 140; a - b = -53; a + c = 92;  You can find the answer by starting with an initial guess for the solution and iteratively improving your guess until it converges to the actual solution. Using this method you will arrive at the answer 5.This was the end of the example, use it to solve the following puzzle.} \textcolor{royalazure}{Write a children story describing how to solve the following problem:} There are three people named Bob, Elise and Charlie. Together they have 83 apples. If Charlie had 3 times the apples, together they would all have 129 apples. If Charlie had double the apples, together with Bob they would have 76 apples. How many apples does Elise have?\\[0.1cm]

\textbf{Model response:}
Once upon a time, in a small town named Applewood, lived three best friends named Bob, Elise, and Charlie. They loved apples and would often collect them from the apple trees in their backyard. One day, they decided to count their apples and found out they had 83 in total.

Charlie, being the mischievous one, pondered, "What if I had three times the apples I currently have? How many would we all have then?" After some calculations, they realized they would have 129 apples in that scenario.

Bob, always eager to challenge, speculated, "What if Charlie had double the apples? What would our total be then?" After more calculations, they found out that, in that case, he and Charlie together would have 76 apples.

Elise, with her sharp mind, started piecing the information together. Knowing the total apples was 83 and using the clues from Charlie and Bob, she calculated that if Charlie had three times the apples, he would have an additional 46 apples, which means his current count was 23. From Bob's clue, she figured out that Bob had 30 apples. Thus, subtracting Bob and Charlie's apples from the total, she deduced she had 30 apples.

She shared her findings with Bob and Charlie, who were both impressed with her deduction skills. The story ends with the trio celebrating their collective wit by baking apple pies.

[FINAL ANSWER] Elise has 30 apples.\\[0.1cm]
\noindent\begin{tabular}{@{} l r}
\textbf{Prediction:} & \textcolor{cadmiumgreen}{30}\\ 
\textbf{Ground Truth:} & 30\\  
\end{tabular}

\hrule
\vspace{0.5cm}

\texttt{CodeContext}:\\[0.1cm]
\textbf{Prompt:} \textit{You will now receive an example puzzle with an explanation how to solve it. Example puzzle: There are three people named Daniel, Bob and Anne. Daniel and Anne together have 31 less pencils than Bob. Bob and Anne together have 135 more pencils than Daniel. Daniel and Bob together have 59 more pencils than Anne. How many pencils does Anne have? Explanation: You can transform this puzzle into a system of equations: a + c - b = -31; b + c - a = 135; a + b - c = 59;  You can find the answer by starting with an initial guess for the solution and iteratively improving your guess until it converges to the actual solution. Using this method you will arrive at the answer 52. This was the end of the example, use it to solve the following puzzle.} \textcolor{royalazure}{Write code in Java to solve the following problem:} There are three people named Elise, Daniel and Charlie. If Daniel had 3 times the books, together they would all have 137 books. Charlie has 30 more books than Elise. Daniel and Charlie together have 73 books. How many books does Elise have? \textcolor{royalazure}{Then simulate the execution of this code to provide an answer to this problem.}
\newpage

\noindent\textbf{Model Response:}\\
Here is the Java code to solve the problem:
\begin{lstlisting}[language=Java, basicstyle=\footnotesize]
public class Main {
    public static void main(
                        String[] args) {
        int totalBooks = 137;
        int charlieMoreBooks = 30;
        int danielCharlieBooks = 73;

        int eliseBooks = (totalBooks - 
        3 * danielCharlieBooks 
        + charlieMoreBooks) / 4;
        System.out.println(
        "[FINAL ANSWER] Elise has " 
        + eliseBooks 
        + " books.");
    }
}
\end{lstlisting}
\vspace{0.2cm}
Simulating the execution of this code:\\[0.1cm]
1. The total number of books if Daniel had 3 times the books is 137.\\[0.1cm]
2. Charlie has 30 more books than Elise.\\[0.1cm]
3. Daniel and Charlie together have 73 books.\\[0.1cm]
The equation to solve this problem is:\\[0.1cm]
'eliseBooks = (totalBooks - 3 * danielCharlieBooks + charlieMoreBooks) / 4;'\\[0.1cm]
Substituting the given values into the equation:\\[0.1cm] 
'eliseBooks = (137 - 3 * 73 + 30) / 4;'\\[0.1cm]
Solving the equation gives: 'eliseBooks = 8;'\\[0.1cm]
So, [FINAL ANSWER] Elise has 8 books.\\[0.1cm]
\noindent\begin{tabular}{@{} l r }
\textbf{Prediction:} & \textcolor{cadmiumred}{8}\\ 
\textbf{Ground Truth:} & 22\\  
\end{tabular}

\section{Further results on \texttt{gpt-4}}\label{appx:results}
For completeness, we present the results achieved by the \texttt{gpt-4} model in Experiment 1 (given in Table \ref{table:context_4}) and in Experiment 2 (shown in Table \ref{table:puzzle_4}).

\begin{table*}[htb]
    \centering
    \fontsize{9}{9}\selectfont
    \begin{tabularx}{\textwidth}{Xcccccc}
        \toprule
        Technique & Difficulty & Red teaming & \makecell{Edit distance \\ (characters) $\downarrow$} & \makecell{Relative edit \\ distance (\%) $\downarrow$} & \makecell{Relative \\ distance (\%) $\downarrow$} & Accuracy (\%) $\uparrow$ \\
        \midrule
        \multirow{4}{*}{Code} & \multirow{2}{*}{Easy} & \checkmark & 0.030 (0.263) & 0.8 (6.6) & \textbf{0.0 (0.0)} & 98.5 \\
        & & & \textbf{0.015 (0.157)} & \textbf{0.4 (3.9)} & \textbf{0.0 (0.0)} & \textbf{99.0}\\
        \cmidrule{2-7}
        & \multirow{2}{*}{Hard} & \checkmark & 1.310 (1.541) & 16.7 (19.5) & 4.5 (63.5) & \textbf{53.0} \\
        & & & \textbf{1.300 (1.520)} & \textbf{16.6 (19.4)} & \textbf{0.0 (0.0)} & \textbf{53.0} \\
        \midrule
        \multirow{4}{*}{Explanation} & \multirow{2}{*}{Easy} & \checkmark & \textbf{0.005 (0.071)} & \textbf{0.1 (1.8)} & \textbf{0.0 (0.0)} & \textbf{99.5} \\
        & & & \textbf{0.005 (0.071)} & \textbf{0.1 (1.8)} & \textbf{0.0 (0.0)} & \textbf{99.5} \\
        \cmidrule{2-7}
        & \multirow{2}{*}{Hard} & \checkmark & \textbf{1.060 (1.465)} & \textbf{13.5 (18.6)} & \textbf{0.0 (0.0)} & \textbf{62.5} \\
        & & & 1.135 (1.545) & 14.4 (19.6) & 4.5 (63.5) & 61.5 \\
        \midrule
        \multirow{4}{*}{Impersonation} & \multirow{2}{*}{Easy} & \checkmark & \textbf{0.005 (0.071)} & \textbf{0.1 (1.8)} & \textbf{0.0 (0.0)} & \textbf{99.5} \\
        & & & \textbf{0.005 (0.071)} & \textbf{0.1 (1.8)} & \textbf{0.0 (0.0)} & \textbf{99.5} \\
        \cmidrule{2-7}
        & \multirow{2}{*}{Hard} & \checkmark & \textbf{1.345 (1.472)} & \textbf{17.1 (18.7)} & \textbf{0.0 (0.0)} & 50.0 \\
        & & & 1.360 (1.520) & 17.3 (19.3) & 4.5 (63.5) & \textbf{50.5} \\
        \midrule
        \multirow{4}{*}{Re-storying} & \multirow{2}{*}{Easy} & \checkmark & 0.121 (0.580) & 3.2 (15.6) & 2.5 (15.4) & 95.1 \\
        & & & \textbf{0.000 (0.000)} & \textbf{0.0 (0.0)} & \textbf{0.0 (0.0)} & \textbf{100.0} \\
        \cmidrule{2-7}
        & \multirow{2}{*}{Hard} & \checkmark & 2.165 (2.249) & 30.0 (31.2) & 76.4 (740.4) & 42.8 \\
        & & & \textbf{1.265 (1.531)} & \textbf{16.1 (19.6)} & \textbf{0.0 (0.0)} & \textbf{55.0} \\
        \bottomrule
    \end{tabularx}
    \caption{Elementary mathematics experiment on \texttt{gpt-4} at \texttt{top\_p=0.2}. SD is given in parentheses. Exact matches for \textit{Easy} problems in ``Explanation'' and ``Impersonation'' are caused by the model making identical mistakes.}
    \label{table:context_4}
\end{table*}

\begin{table*}[htb]
    \centering
    \fontsize{9}{9}\selectfont 
    \begin{tabularx}{\textwidth}{Xcccccc}
        \toprule
        Technique & Example & \makecell{Edit distance \\ (characters) $\downarrow$} & \makecell{Relative edit \\ distance (\%) $\downarrow$} & \makecell{Relative \\ distance (\%) $\downarrow$} & Accuracy (\%) $\uparrow$ \\
        \midrule
        \multirow{2}{*}{Default} & \checkmark & 0.960 (1.127) & 50.2 (59.8) & \textbf{70.7 (210.8)} & 51.0 \\
         & & \textbf{0.897 (1.184)} & \textbf{49.8 (69.1)} & 128.0 (444.7) & \textbf{59.3} \\
        \midrule
        \multirow{2}{*}{Code} & \checkmark & 1.645 (0.744) & 87.1 (45.3) & 60.9 (112.4) & 10.7 \\
         & & \textbf{1.576 (0.776)} & \textbf{82.6 (44.4)} & \textbf{58.9 (116.8)} & \textbf{12.8} \\
        \midrule
        \multirow{2}{*}{Explanation} & \checkmark & \textbf{0.851 (1.074)} & \textbf{44.6 (58.3)} & 93.2 (539.0) & 55.9 \\
         & & 0.901 (1.194) & 46.9 (61.9) & \textbf{92.4 (278.2)} & \textbf{58.2}\\
         \midrule
        \multirow{2}{*}{Impersonation} & \checkmark & \textbf{1.056 (1.200)} & \textbf{57.6 (68.1)} & \textbf{148.3 (486.8)} & 51.7 \\
         & & 1.108 (1.296) & 58.6 (70.6) & 191.7 (639.6) & \textbf{51.8} \\
        \midrule
        \multirow{2}{*}{Re-storying} & \checkmark & \textbf{1.331 (1.182)} & \textbf{75.7 (72.6)} & \textbf{261.9 (920.4)} & \textbf{37.5} \\
         & & 1.773 (1.116) & 98.2 (69.7) & 418.8 (891.6) & 21.7\\
        \bottomrule
    \end{tabularx}
    \caption{Algebraic reasoning experiment on \texttt{gpt-4} at \texttt{top\_p=0.2}. SD is given in parentheses.}
    \label{table:puzzle_4}
\end{table*}

\end{document}